%% file: FashionTex.tex
\documentclass[sigconf]{acmart}

\copyrightyear{2023}
\acmYear{2023}
\setcopyright{acmlicensed}\acmConference[SIGGRAPH '23 Conference Proceedings]{Special Interest Group on Computer Graphics and Interactive Techniques Conference Conference Proceedings}{August 6--10, 2023}{Los Angeles, CA, USA}
\acmBooktitle{Special Interest Group on Computer Graphics and Interactive Techniques Conference Conference Proceedings (SIGGRAPH '23 Conference Proceedings), August 6--10, 2023, Los Angeles, CA, USA}
\acmPrice{15.00}
\acmDOI{10.1145/3588432.3591568}
\acmISBN{979-8-4007-0159-7/23/08}

\AtBeginDocument{%
  \providecommand\BibTeX{{%
    \normalfont B\kern-0.5em{\scshape i\kern-0.25em b}\kern-0.8em\TeX}}}

\begin{document}

\title{FashionTex: Controllable Virtual Try-on with Text and Texture}

\author{Anran Lin}
\email{anranlin@link.cuhk.edu.cn}
\orcid{0009-0006-2550-598X}
\affiliation{%
  \institution{SSE, CUHKSZ}
  \country{China}
}

\author{Nanxuan Zhao}
\email{nanxuanzhao@gmail.com}
\affiliation{%
  \institution{Adobe Research}
  \country{USA}
}

\author{Shuliang Ning}
\email{shuliangning@link.cuhk.edu.cn}
\affiliation{%
  \institution{FNii, CUHKSZ}
  \institution{SSE, CUHKSZ}
  \country{China}
}

\author{Yuda Qiu}
\email{yudaqiu@link.cuhk.edu.cn}
\affiliation{%
  \institution{FNii, CUHKSZ}
  \institution{SSE, CUHKSZ}
  \country{China}
}

\author{Baoyuan Wang}
\email{zjuwby@gmail.com}
\affiliation{%
  \institution{Xiaobing.AI}
  \country{China}}

\author{Xiaoguang Han}
\authornote{Corresponding author.}
\email{hanxiaoguang@cuhk.edu.cn}
\affiliation{%
  \institution{SSE, CUHKSZ}
  \institution{FNii, CUHKSZ}
  \country{China}
}


\begin{abstract}
  Virtual try-on attracts increasing research attention as a promising way for enhancing the user experience for online cloth shopping. Though existing methods can generate impressive results, users need to provide a well-designed reference image containing the target fashion clothes that often do not exist. To support user-friendly fashion customization in full-body portraits, we propose a multi-modal interactive setting by combining the advantages of both text and texture for multi-level fashion manipulation. With the carefully designed fashion editing module and loss functions, FashionTex framework can semantically control cloth types and local texture patterns without annotated pairwise training data. We further introduce an ID recovery module to maintain the identity of input portrait. Extensive experiments have demonstrated the effectiveness of our proposed pipeline. Code for this paper are at \url{https://github.com/picksh/FashionTex}.
\end{abstract}

\ccsdesc[500]{Computing methodologies~Computer vision}
\ccsdesc[500]{Computing methodologies~Image manipulation}

\keywords{image manipulation, controllable fashion generation, multi-modal learning }

\begin{teaserfigure}
  \includegraphics[width=\textwidth]{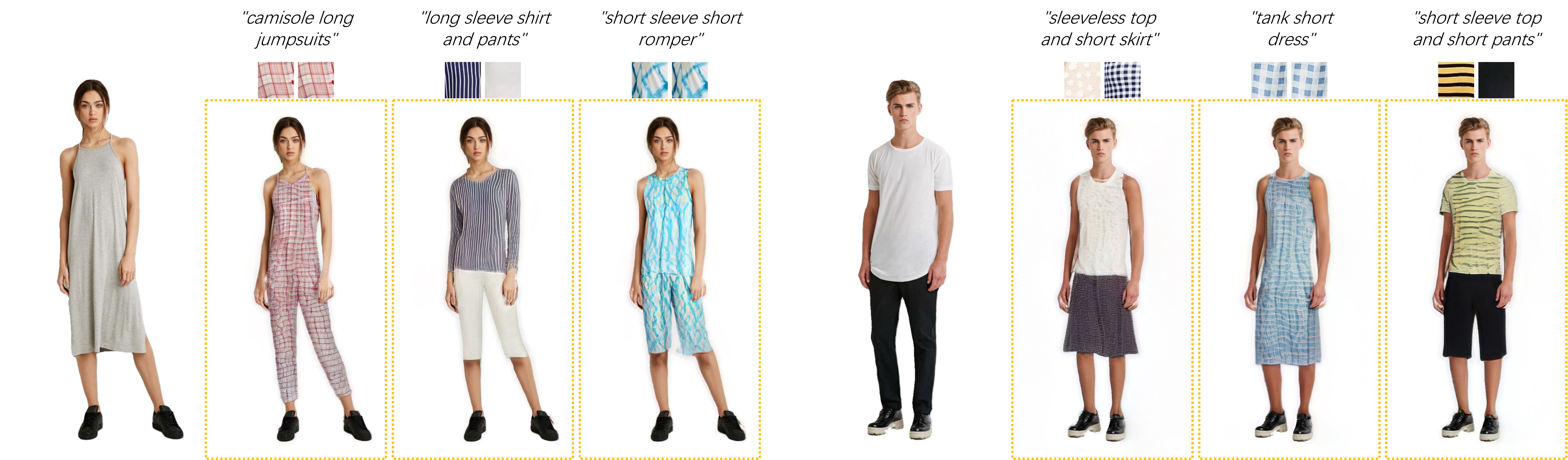}
  \caption{FashionTex performs the full-body virtual try-on with multi-modal
controls over garment type and texture pattern, allowing anyone to design personalized clothes with simple interactions. We show two design cases here. For each of the cases, the input portrait is presented on the left, with three outputs under different conditions. Each condition contains one text prompt and two texture patches for upper and lower cloth parts.}
  \label{fig:teaser}
\end{teaserfigure}


\maketitle

\input{Sections/1_intro.tex}
\input{Sections/2_related.tex}
\input{Sections/3_method.tex}

\input{Sections/4_exp.tex}
\input{Sections/5_conclusion.tex}

\begin{acks}
The work was supported in part by NSFC with Grant No. 62293482, the Basic Research Project No. HZQB-KCZYZ-2021067 of Hetao Shenzhen-HK S\&T Cooperation Zone, the National Key R\&D Program of China with grant No. 2018YFB1800800, the Shenzhen Outstanding Talents Training Fund 202002, the Guangdong Research Projects No. 2017ZT07X152 and No. 2019CX01X104, the Guangdong Provincial Key Laboratory of Future Networks of Intelligence (Grant No. 2022B1212010001), and the Shenzhen Key Laboratory of Big Data and Artificial Intelligence (Grant No. ZDSYS201707251409055). It was also supported in part by  Outstanding Yound Fund of Guangdong Province with No.  2023B1515020055. It was also sponsored by CCF-Tencent Open Research Fund.
\end{acks}
\bibliographystyle{ACM-Reference-Format}
\bibliography{sample-base}


\end{document}

%% file: Sections/1_intro.tex
\section{Introduction}
\label{sec:intro}
Since the advent of e-commerce, the popularity of online shopping grows sharply. As of 2021, Amazon has over 200 million Prime subscribers~\cite{amazon}. To enhance user experience for online cloth shopping, virtual try-on has emerged to allow customers to try out the products before buying. It attracts increasing research attention and numerous methods~\cite{xie2021towards,lewis2021tryongan,neuberger2020image,raj2018swapnet, lee2022high} have been proposed. These works transfer the fashion (\textit{i.e.}, clothes) from a reference image with an existing product to the target one. Though they can generate impressive results, users need to provide reference images containing target fashion clothes that often do not exist.



We thus seek an interactive way of supporting user-friendly fashion customization. Inspired by the recent success of text-based image manipulation~\cite{patashnik2021styleclip,kwon2022clipstyler,wei2022hairclip,kim2022diffusionclip,xu2022predict,couairon2022flexit} powered by visual-textual pretrained models (\textit{e.g.}, CLIP \cite{radford2021learning}), we also aim to conduct text-based virtual try-on, which is natural to everyone. In practical usage, we found that text is efficient for controlling high-level semantic changes (\textit{i.e.}, cloth types), but fails to alter local details (\textit{i.e.}, textures). To solve this problem, we introduce a new interactive setting by combining the advantages of both text and texture for virtual try-on. More specifically, given a full-body portrait, users can edit cloth types (\textit{e.g.}, long sleeves and short pants ) through texts and local patterns on clothes through texture patches, as shown in Fig.~\ref{fig:teaser}. Besides, we build our model on StyleGAN~\cite{karras2019style}, following the recent common practice\cite{sarkar2021style, wei2022hairclip}.



However, our task has three unique challenges. First, fashions with full-body portraits display diverse poses, cloth types, and appearances. While text takes charge of global structure deformation, texture patches target on changing local appearance. How to learn a model to precisely edit specified regions in different levels without modifying characteristics of the original body (\textit{e.g.}, face, skin, and pose) remains difficult. Second, collecting a large dataset of pairwise data (\textit{i.e.}, original portrait with text instruction, and modified portrait) is impractical. How to enable a model for understanding textual input is unknown. Third, when employing StyleGAN on real images, the reconstruction error often occurs, which is unacceptable for our task.



To this end, we propose a novel pipeline called \textit{Fashion\textbf{Tex}}, controlling virtual try-on with \textbf{Tex}t and \textbf{Tex}ture. We first explore the latent codes of StyleGAN by grouping them into different levels, disentangling textures from structures for better control. A fashion editing module is then designed to learn two different mappers for textual and textural inputs based on disentangled latent codes respectively. To deform the cloth types based on the input text without paired training data, we utilize the CLIP embedding by proposing a new type loss. Our type loss can accurately modify the cloth regions (\textit{e.g.}, only change the sleeves from long to short) without influencing the surrounding parts. As our model relies on a perfect inversion, the errors that happened during the reconstruction of real images lead to the loss of portrait identity. To deal with this problem, we present an ID recovery module for restoring the input identity to obtain acceptable results.

Through extensive experiments, we have demonstrated the effectiveness of our proposed pipeline.
Our main contributions are:

\begin{itemize}
    \item To the best of our knowledge, we are the first attempt to conduct the full-body virtual try-on with multi-modal controls~(\textit{i.e.}, text and texture patches), allowing anyone to design personalized clothes with simple interactions.
    \item We propose a fashion editing module for better disentangling the editings between textual and textural inputs, and a novel CLIP-based type loss for accurately adjusting the cloth types without paired training data.
    \item We introduce an ID recovery module to mitigate the reconstruction errors caused by StyleGAN inversion, and obtain satisfied results on real images.
    
\end{itemize}

%% file: Sections/2_related.tex
\section{Related work}
\label{sec:relat}
\paragraph{Portrait Image Generation.}
The Generative Adversarial Networks (GAN) \cite{goodfellow2020generative} has been widely leveraged for image generation in recent years \cite{pix2pix2017, CycleGAN2017, karras2017progressive, karras2020analyzing}. Its follow-up work StyleGAN\cite{karras2019style} can further generate high-resolution photo-realistic images while maintaining a disentangled embedding, inspiring recent works on human-centric image synthesis works~\cite{sarkar2021style,albahar2021pose,fruhstuck2022insetgan,fu2022stylegan}.

InsetGAN\cite{fruhstuck2022insetgan} combines different parts from multiple pretrained GANs to obtain a photo-realistic full-body image. StyleGAN-Human\cite{fu2022stylegan} offers editing benchmark on clothed full-body images with prior conditions. 
It adopts facial editing methods\cite{shen2020interfacegan, wu2021stylespace, shen2021closed} that use preset StyleGAN latent space direction to change the appearance.
Although these methods \cite{fruhstuck2022insetgan,fu2022stylegan} are well-designed for photo-realistic human synthesis, they can only generate random images and fail to accomplish image synthesis according to specific conditions, which is a crucial
requirement for the virtual try-on scenario.

\vspace{-2mm}
\paragraph{Image-based Fashion Editing.}
Given a portrait image and user instructions, such as a reference fashion image \cite{han2018viton, wang2018toward, xie2021vton, yu2019vtnfp, raj2018swapnet}  or a texture style \cite{Brown22,xian2018texturegan,issenhuth2021edibert,albahar2019guided}, this kind of methods aims to synthesize the target image sharing the same identity as the input portrait while wearing the specific fashion. Most existing works rely on a well-designed fashion product image. Some works~\cite{xie2021towards,sarkar2020neural} are designed to exchange garments between two images with different portraits, without the requirement of a product reference image. Nevertheless, this process still needs a user-satisfied fashion garment, which may be hard to create or provide.
Another kind of work \cite{gunel2018language,zhu2017your} is proposed to manipulate fashion images based on the text description. 
More specifically, a few existing works try to conduct fashion editing based on a user-specified attribute \cite{zhu2016generative, ak2019attribute,chen2020tailorgan}, such as \textit{sleeve length, color and pattern}. 
However, only relying on text descriptions may be hard to control local details. Instead, our work proposes to use a multi-modal control by complementing text inputs with texture patches.

\vspace{-2mm}
\paragraph{Clip-based Image Editing.}
Recently, Contrastive Language–Image Pre-training (CLIP)\cite{radford2021learning} has shown great power in multimodal learning. Benefit from the vision-language semantic alignment, the combination of the conditional generative model and CLIP brings massive amazing results\cite{ramesh2022hierarchical, rombach2021highresolution}. DiffusionCLIP\cite{kim2022diffusionclip} uses diffusion model\cite{zhu2016generative} with CLIP, achieving high-quality zero-shot image manipulation results. FlexIT\cite{couairon2022flexit} embeds the image and text with CLIP encoders to find the target point and edits in the latent space of the VQ-GAN autoencoder \cite{esser2021taming,yu2021vector}. StyleCLIP\cite{patashnik2021styleclip} combines the powerful image synthesis ability of StyleGAN and the amazing image-text representation ability of CLIP, showing high-quality results of human face editing. HairCLIP\cite{wei2022hairclip} then introduces the idea of hair editing with a CLIP-based unified architecture for text and image reference. However, the above methods cannot directly apply to full-body fashion images because of lacking control over complex poses, types, and cloth texture. In this work, we propose a new CLIP-based type loss that can better capture the difference between the original image and the target one to perform global type editing and subtle attribute changes.

%% file: Sections/3_method.tex
\section{Methodology}
\label{sec:method}
\input{Figs/Pipeline.tex}

Given a full-body portrait $I_i$, FashionTex aims to edit the fashion clothes for trying on the original body, by using text prompts $t$ to indicate changes in cloth types and reference RGB patches, $P=\{p_{up},p_{low}\}$, for the textural pattern of upper and lower clothes. Inspired by previous works\cite{wei2022hairclip, fruhstuck2022insetgan}, we take advantage of the generation ability of a pretrained StyleGAN on human bodies\cite{fu2022stylegan}. As shown in Fig.~\ref{fig:pipeline}(a), FashionTex first inverts the input portrait image $I_i$ back into the latent $W+$ space of StyleGAN using the e4e encoder~\cite{tov2021designing}. By manipulating this latent vector $w$, we can obtain the edited fashion design $I_e$ from the pretrained StyleGAN $G_{H}$, with the new latent vector $w'$ based on the input text $t$ and texture patches $P$. We design a fashion editing module for predicting an offset $\Delta w$, and compute the edited latent code as $w'=w+\Delta w$ (Sec.~\ref{sec:editing}). The final try-on image $I_o$ is derived by fusing the input portrait $I_i$ with the edited fashion design $I_e$ (Sec.~\ref{sec:real_image}). Here we explain our method in more detail.

\input{Sections/3.1_method_editing}
\input{Sections/3.2_method_realimage}

%% file: Figs/Pipeline.tex
\begin{figure*}[t]
    \vspace{-1mm}
  \centering
   \includegraphics[width=\linewidth]{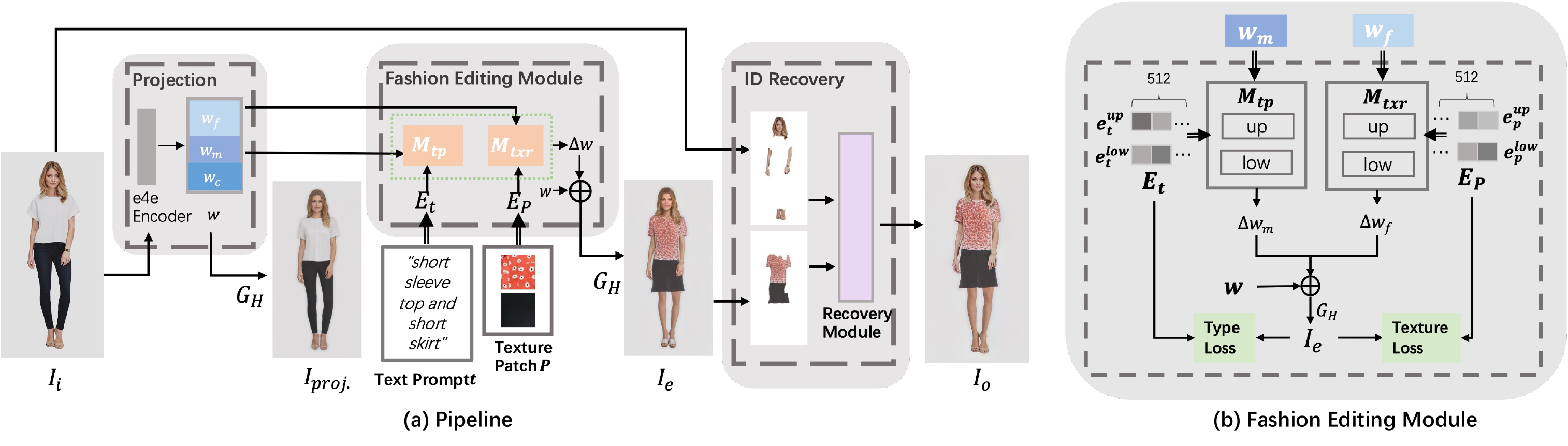}

    \vspace{-1mm}
   \caption{(a) Overview of our pipeline. Our framework contains three modules: latent code projection, fashion editing, and ID recovery modules. In projection, we use e4e encoder to invert input image $I_{i}$ to latent code $w=[w_c, w_m, w_f]$. $I_{proj.}$ is the reconstruction result from $w$ using StyleGAN-Human generator $G_H$. In fashion editing module, we use two mappers to handle type and texture editing. We feed part-aware editing modules with text embedding $E_t$ and texture patch embedding $E_p$ to produce the offsets, $\Delta w_m$, and $\Delta w_f$ that lead to attribute changes in StyleGAN-Human latent space. The edited fashion result $I_{e}$ is generated from $(w+\Delta w)$ using $G_H$ and will further gone through an ID recovery module to obtain the final output image $I_{o}$, maintaining the human characteristics of the input $I_{i}$. (b) The fashion editing module.} 
   \label{fig:pipeline}
   \vspace{-3mm}
\end{figure*}

%% file: Sections/3.1_method_editing.tex
\subsection{Fashion Clothes Manipulation}
\label{sec:editing}
\paragraph{Editing with StyleGAN latent code.} Our model takes multi-modal interactions as conditions, controlling different levels of fashion clothes. The text takes charge of high-level semantic structures, such as sleeve length and neckline shape, while texture tends to modify low-level local patterns. To incorporate these differences in a single model, we seek the help of well pre-trained StyleGAN\cite{fu2022stylegan} given its wide usage in high-quality and realistic synthesis tasks. Different layers in StyleGAN controls different levels of detail in the generated image, disentangling the appearance from structure. Following this idea, we first project the input image $I_i$ into a StyleGAN latent code $w$, and divide the latent code into coarse, medium, and fine groups as $w=[w_c, w_m, w_f]$. The objective is to use $w_m$ control the garment type mainly with shapes, and $w_f$ to edit the texture and color details, relating more to fine-grain features. Following the common practice\cite{wei2022hairclip, fu2022stylegan}, we examine the interpretability of the corresponding latent space by style mixing. Given a source and reference images pair, we copy each layer from the latent code of the reference to those of the source and evaluate the changes in the generated image, comparing with the source portrait. In particular, we group our latent space as 1$\sim$4 for $w_c$, 5$\sim$8 for $w_m$, 9$\sim$18 for $w_f$.
\vspace{-2mm}
\paragraph{Fashion editing module.} We then design a fashion editing module for predicting the updated latent code $w'$ conditioned on text prompt $t$ and texture patch $P$, as shown in Fig.~\ref{fig:pipeline}(b). Instead of directly predicting the final code $w'$, we aim to predict an offset $\Delta w$ for each condition with more precise control, \begin{equation}\label{eq:delta_wm}
\vspace{-1mm}
\begin{aligned}
    w'&=[w_c, w_m+\Delta w_m, w_f+\Delta w_f].
\end{aligned}
\end{equation}
For input text condition, we utilize recent powerful joint representations of Contrastive Language-Image Pre-training(CLIP)\cite{radford2021learning}  to encode $t$ as a text embedding $E_t$. CLIP has successfully been used in many visual-text tasks. For input texture patch, we use a pretrained VGG Network to capture the rich variation in texture patterns, following \cite{men2020controllable}, and obtain the texture embedding as $E_p$. With these two embeddings, we build two different mappers (\textit{i.e.}, type mapper $M_{tp}$ and texture mapper $M_{txr}$) to learn the manipulation respectively by fusing input latent codes and conditions together as $\Delta w_m=M_{tp}(w_m, E_t)$ and $\Delta w_f=M_{txr}(w_f,E_p)$.

When designing our model, because of the unique feature of fashion, we find that the distributions of upper and lower fashion clothes are quite different. For example, we observe that the upper part of clothes often have more subtle changes in shapes, such as long sleeve versus short sleeve and round neck versus v-neck, while the lower part may have changes in structure, such as changing pants to a skirt.
To better decoupling these two parts, we separately modeling them with two part-aware conversion modules in each mapper. Each part conversion module consists of a modulation module to capture the condition-based guidance. The structure of the modulation module is derived from \cite{wei2022hairclip,tan2021efficient, park2019semantic,huang2017arbitrary}.  Taking the type branch (\textit{i.e.}~text prompt) as an example, we split the input text 
before sending into the CLIP for obtaining $E_t=\{e_{t}^{up}, e_{t}^{low}\}$. After fusing them separately with $w_m$, the output feature $\Delta w_m^{up}$ and $\Delta w_m^{low}$ would be added together to get the offset latent code $\Delta w_m$ for the whole-body type adjustment. More specifically, the text embedding $e_{t}^{up}$ and $e_t^{low}$ are first fed into two fully-connected layers to obtain $\gamma^{up}, \beta^{up}$ and $\gamma^{low}, \beta^{low}$, respectively. The complete process can be formulated as follows:
\begin{equation}
\vspace{-2mm}
\begin{aligned}
\Delta w_m^{i}&=\beta^i + \gamma^i \frac{w_m^{i}-\mu_{w_m^{i}}}{\sigma_{w_m^{i}}} , i\in{[up, low]}\\
    \Delta w_m &= \sum_{i} \Delta w_m^{i}
    \label{eq:modulation}
\end{aligned}
\vspace{-1mm}
\end{equation}
where $\mu_{w_m^{i}}$ is the mean of $w_m^{i}$ and $\sigma_{w_m^{i}}$ is the variance. 
Similarly, we follow the same process to obtain the offset $\Delta w_f$  for the fashion texture.

\subsection{Loss Functions}
One of the biggest challenges of our task is training without annotated dataset as collecting large-scale pairwise data containing condition inputs and portrait images is impractical. To make the model trainable, we design a novel type loss, allowing the model to precisely adjust the cloth types without touching the remaining characteristics of the input portrait, such as face, skin, and pose. To achieve high-quality results, we also introduce other auxiliary losses, which will be introduced next.

\vspace{-2mm}
\paragraph{Type Loss.} 
Given the powerful visual-text representation power, one may think that a straightforward way is to directly calculate the cosine distance between the edited fashion design $I_e$ and text prompts $t$. However, this does not work in our case as directly calculating the loss in a global manner will lead the model to overlook details. 
Another possible solution is to use a mask to help the model concentrate on the modifiable regions, but this limits the recognition power of the original CLIP embedding. To tackle the above problem, we exploit the linear operations supported by CLIP embeddings~\cite{jia2021scaling,couairon2022flexit}. Based on the cloth tags/types $t_i$ label in original training data, we can compute a latent code for the unmodifiable region by subtracting the CLIP embedding of $t_i$ from the CLIP embedding of input portrait image $I_i$, as $E_{I_{un}}=E_{I_i}-E_{t_i}$. This is benefited from the well aligned embedding space of CLIP for text and image modalities. Then, we can obtain a calibrated ground truth CLIP embedding by adding this unmodifiable embedding on to the CLIP embedding of target text prompt $E_t$ as $\widetilde{E_{t}} = E_{I_{un}} + E_t$. Thus, we can compute the difference between this calibrated ground truth and our obtained results in a more accurate manner with:
\begin{equation} 
\label{eq:typeloss}
    L_{type}=1-cos(E_{I_{e}}, \widetilde{E_{t}}).
\vspace{-1mm}
\end{equation}

\vspace{-2mm}
\paragraph{Texture Loss.}
We design a texture loss to transfer the texture of input patch to the edited fashion clothes for the upper and lower parts, respectively.To emphasize the local spatial pattern in the reference texture $P$, we compute the feature correlations for RGB texture patches. More specifically, we first obtain the feature maps for an image patch by pretrained VGG-19\cite{simonyan2014very}, and extract the outputs for the last four layers, which are more relevant to the pixel-level characteristics. For the feature map $F_{i}$ of each layer, we calculate the correlations by the Gram matrix\cite{portilla2000parametric}, \textit{i.e.} $G_i = F_{i}F_{i}^{T}$,
\begin{equation}
\vspace{-1mm}
    L_{txr}=\sum_{i=1}^{4}\lVert G_{i}(I_{e}^{crop}),G_{i}(P) \rVert _1,
\label{eq:cliploss}
\end{equation}
where $I_{e}^{crop}$ is a random cropped patch from the corresponding semantic region of $I_{e}$. For example, when the textual condition is "pants", the patch $I_{e}^{crop}$ is fetched from the lower region of $I_{e}$ according to the human parsing results $\mathcal{P}$ from \cite{ak2019semantically}.

\vspace{-1mm}
\paragraph{Reconstruction Loss.} 
To better preserve the unchangeable regions, we further add a set of reconstruction losses for preserving identity $L_{id}$, background (\textit{i.e.}, other non-cloth regions) $L_{bg}$ and skin color $L_{skin}$.

\textit{Identity loss.} We rely on the pretrained ArcFace network \cite{deng2019arcface} $Arc(\cdot)$ to keep the face identity by calculating the cosine distance between the original and eddited image in ArcFace embedding space:
\begin{equation}
    L_{id}=1-cos(Arc(I_{e}),Arc(I_{i})).
    \label{eq:id_loss}
\end{equation}

To obtain the background region, we use the human parsing results $\mathcal{P}$ in binary format generated from \cite{rao2022denseclip} by removing the cloth region. We  then obtain the background loss to preserve the non-cloth region by calculating the $L2$ distance:
\begin{equation}\label{eq:bgloss}
\begin{aligned}
  L_{bg}=\lVert(I_{e}*\mathcal{P}_{bg}(I_{e})
  -I_{i}*\mathcal{P}_{bg}(I_{i}))\rVert _2.
\end{aligned}
\end{equation}
\textit{Skin color loss.} Though background loss has constrained on the skin also, we find that further add a loss to help better preserve the skin color is necessary. Similar to the background, we obtain the skin parsing binary mask as $\mathcal{P}_{skin}$. We then convert the colors within this region into LAB color space which is more aligned with human perception. The skin color loss is to constrain the average color changes in the corresponding skin region with $L1$ as:
\begin{equation}\label{eq:skinloss}
\begin{aligned}
  L_{skin}=\lVert(&Avg(Lab(I_{e})*\mathcal{P}_{skin}(I_{e}))
-\\& Avg(I_i*\mathcal{P}_{skin}(I_i)))\rVert_1
\end{aligned}
\vspace{-1mm}
\end{equation}

\vspace{-2mm}
\paragraph{Regularization Loss.} We also apply an L2 regularization loss on $\Delta w$ to enable stable training without generating too large offsets as:
\begin{equation}
\vspace{-2mm}
    L_{norm}=\left\|\Delta w \right \| _2.
    \label{eq:l2norm}
\vspace{1mm}
\end{equation}
In summary, the final loss for training FashionTex to generate eddited fashion image is:
\begin{equation} \label{eq:totalloss}
    \begin{aligned}
    L&=\lambda_{type} L_{type} + \lambda_{txr} L_{txr} +\lambda_{id} L_{id} + \lambda_{skin} L_{skin} \\
    &+\lambda_{bg} L_{bg} + \lambda_{norm} L_{norm}, \\
    \end{aligned}
\end{equation} where $\{\lambda_i\}$ are weighted parameters.

%% file: Sections/3.2_method_realimage.tex
\subsection{Identity (ID) Recovery Module}
\label{sec:real_image}
%
As mentioned at the start of this section our model relies on the inversion model for converting the input portrait images into the StyleGAN editable latent codes. However, we find that the inversion methods often struggle with the trade-off between the ability of editing and reconstruction. Especially our fashion images share more diverse appearances, causing reconstruction errors noticeable even with the state-of-the-art inversion methods. We follow ~\cite{fu2022stylegan, tzaban2022stitch} use PTI inversion~\cite{roich2022pivotal} to obtain the latent codes. since PTI inversion needs to finetune the generator for each of the image for obtaining good result, it fails to directly output satisfied results for our task, especially on identities of portraits. A simple answer may be copying the edited fashion cloth region back to the portrait with a guided semantic binary mask. Unfortunately, this does not work as our type changes often adjust the shape of clothes. For example, when adjusting the sleeves to be short, pasting back to the portrait image can generate serious artifacts with some parts of the original sleeves remaining.


To alleviate the identity loss in our final results, while maintaining the well-modified fashion cloths based on input conditions, we use the regularization ability of the StyleGAN space to compensate for the artifacts in part fusion. We design a semantic-aware ID recovery module to obtain satisfied results. In particular, we first fuse the clothes regions in the edited image $I_e$ using a binary semantic mask~\cite{rao2022denseclip} to gain a guided image $I'_e = \mathcal{P}_{cloth}(I_{e})*I_{e}+\mathcal{P}_{bg}(I_{e})*I_{i}$. We then finetune the StyleGAN-Human generator similar to the PTI inversion guiding with LPIPS perceptual~\cite{zhang2018unreasonable} and $L_2$ distances to get the refined output $I_o$:
\begin{equation}\label{eq:recover}
\begin{aligned}
 L_{ID}&=L_{lpips}(I'_{e},I_{o})+ \lVert \mathcal{P}_{bg}(I_{o})*(I_{i}-I_{o}) \rVert _2.
\end{aligned}
\end{equation}
To obtain $I_{o}=G_{H}(I'_{e};\theta^{*})$, we define the optimization as:
\begin{equation}
    \theta^{*} = \mathop{\arg\max}\limits_{\theta^{*}} L_{ID},
    \vspace{-2mm}
\end{equation}
where $\theta^{*}$ is the parameters set of the generator $G_H$.

%% file: Sections/4_exp.tex
\section{Experiments}
\label{sec:expri}
\subsection{Implementation Details}
We train and evaluate our method on Deepfashion-MultiModal dataset\cite{liuLQWTcvpr16DeepFashion, jiang2022text2human}. 
It contains 12,701 full-body human images with human parsing labels of 24 classes and descriptions of each image. 
But in this work, there is no need for segmentation maps or pair-wise captions, we only use the cloth-type labels. 
We first use the full-body image alignment method with the mean body midpoint mentioned in \cite{fu2022stylegan} to process the images and abandon the samples with bad alignment results. For the remaining data, we randomly split 11,265 and 1,136 data for the train and test sets, respectively. 
For the text prompts, we borrow the common practice in online clothing stores, which clusters the garments in attribute combinations, \textit{e.g.} "sleeveless top, and short skirt".
For the reference texture patch, we crop texture patches from datasets \cite{liuLQWTcvpr16DeepFashion, jiang2022text2human} to get realistic clothing textures. 

We utilize pretrained StyleGAN2 model\cite{fu2022stylegan,sarkar2021style} as the generator and pretrained e4e encoder\cite{tov2021designing} as the image encoder to invert images into StyleGAN's latent codes.
The dimensions of the latent code are $18*512$. We keep the code from the coarse layer of StyleGAN unchanged and only perform editing on medium and fine layers.

\input{Figs/result}
\subsection{Results of Multi-modal-guided Fashion Editing}
We show our results on multi-modal fashion editing in Fig.~\ref{fig:results}. As we are the first to work on this new task, there are no available methods for direct comparisons. We thus show our complete results here and leave the evaluations on the individual modality in the following subsections. As can be seen from Fig.~\ref{fig:results}, our FashionTex can deal with various input text prompts and texture patches. The input text prompts can vary from general high-level descriptions "skirt" to fine-grained ones such as "camisole dress" without any pairwise annotated training data. Besides, our model can preserve the input identity and pose well for achieving satisfying try-on results. Even for some less frequently seen design cases (\textit{e.g.}, the $3^{rd}$ output of the $1^{st}$ case in the $2^{nd}$ row.), our model can generate reasonable results. An interesting finding is that FashionTex can automatically find the more suitable cloth even for the same input to meet common sense. For example, as shown in the $2^{nd}$ output of the $2^{nd}$ case each row, it generates long tight joggers for the lady (\textit{i.e.}, upper row), while generating short loose joggers for the man (\textit{i.e.}, lower row). These findings support the effectiveness of our model, and we show more detailed evaluations below.



\input{Sections/4.1_exp_type.tex}

\input{Sections/4.2_exp_texture.tex}

\input{Sections/4.3_ablation.tex}
\vspace{-2mm}


%% file: Figs/result.tex
\begin{figure*}[ht!]
    \vspace{-1mm}
  \centering
  \includegraphics[width=\linewidth]{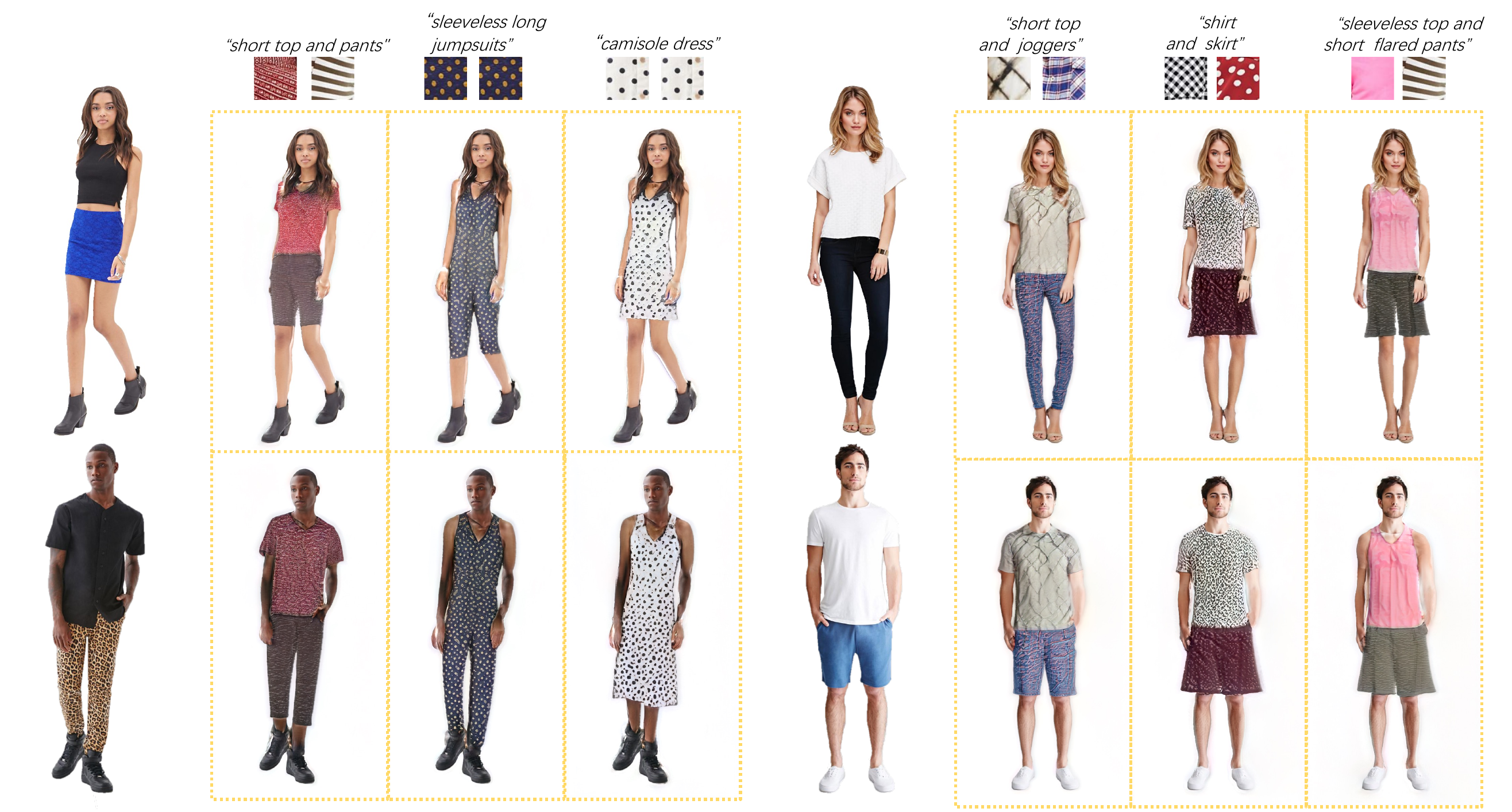}
    \vspace{-3mm}
   \caption{Our results on multi-modal editing on fashion clothes try-on. With simple interaction by text prompts and texture patches, our model can generate try-on results meeting the requirement while keeping the input portrait identity.}
   
   \label{fig:results}
   \vspace{-1mm}
\end{figure*}

%% file: Sections/4.1_exp_type.tex
\subsection{Comparisons on Fashion Type Editing}

\paragraph{Baseline methods.}
To verify the effectiveness of our model on cloth type editing, we compare our method with two state-of-art text-driven image manipulation works. (1) TediGAN \cite{xia2021tedigan} proposes a
visual-linguistic similarity module to project the image and
text into a common embedding space.
After inverting a fashion portrait into the joint latent space, it can achieve the type editing by changing the text embedding for cloth. We follow the official CLIP-based implementation and the default settings for training. (2) StyleCLIP\cite{patashnik2021styleclip} proposes three latent mappers to control the different latent group (\textit{i.e.} $\{w_c, w_m, w_f\}$) in the $W+$ space of StyleGAN, conditioned on a textual description and a source image. We follow the official instruction to train StyleClip, except that we only apply the mapper for the medium layer $w_m$ to perform the type editing task. For each text input, StyleClip needs to train corresponding mappers to perform the editing. As for FashionTex, we only use the type mapper $M_{tp}$ with $w_m$ and remove the proposed ID recovery module for fair comparison since these two methods do not have abilities to recover identities.

\vspace{-2mm}
\paragraph{Metrics.}
To evaluate the performance of each method, we use two metrics: 1) \textit{Accuracy}. It measures whether the model succeeds in getting the target cloth type. We use a human parsing network\cite{rao2022denseclip} pretrained on \cite{liuLQWTcvpr16DeepFashion} to find if the target cloth type is in the manipulated image. Since the parsing results can only reflect the category of clothes (\textit{e.g.}, skirt, pants) and cannot identify more specific attributes (\textit{e.g.}, sleeve length ), we choose four common cloth categories for evaluation, \textit{i.e.}, skirt, pants, dress, and rompers. 2) \textit{FID}. It measures the realism of the generated images by computing the Wasserstein-2 distance between distributions of the generated images and the corresponding type of images in our dataset.
\paragraph{Qualitative Results.}
\input{Figs/comp-text}

The visual comparisons for type editing are shown in Fig.~\ref{fig:com_tex}. When the target cloth type is close to the source image, \textit{e.g.} from "shirt" to "polo shirt", all methods can achieve reasonable results. However, when there are large changes in cloth structure, \textit{e.g.} from "pants" to "skirt", both TediGAN and StyleCLIP struggle to change the original cloth type, while our FashionTex generates the target fashion style as the textual condition. Since our clip loss can pay more attention to the area that needs to be edited, we achieve good results on both subtle attribute transformations and large type changes. Our results have advantages in the preservation of both color and facial information using our reconstruction and regularization losses. 

\vspace{-1mm}
\paragraph{Quantitative Results.} 

\input{Tables/quant.tex}

We show the quantitative results in Tab.~\ref{tab:comparison}, and our model outperforms the previous works for a large margin on both metrics, further demonstrating the advantage of our model on fashion type editing. 


%% file: Figs/comp-text.tex
\begin{figure*}[h]
    \vspace{-2mm}
  \centering
  \includegraphics[width=\linewidth]{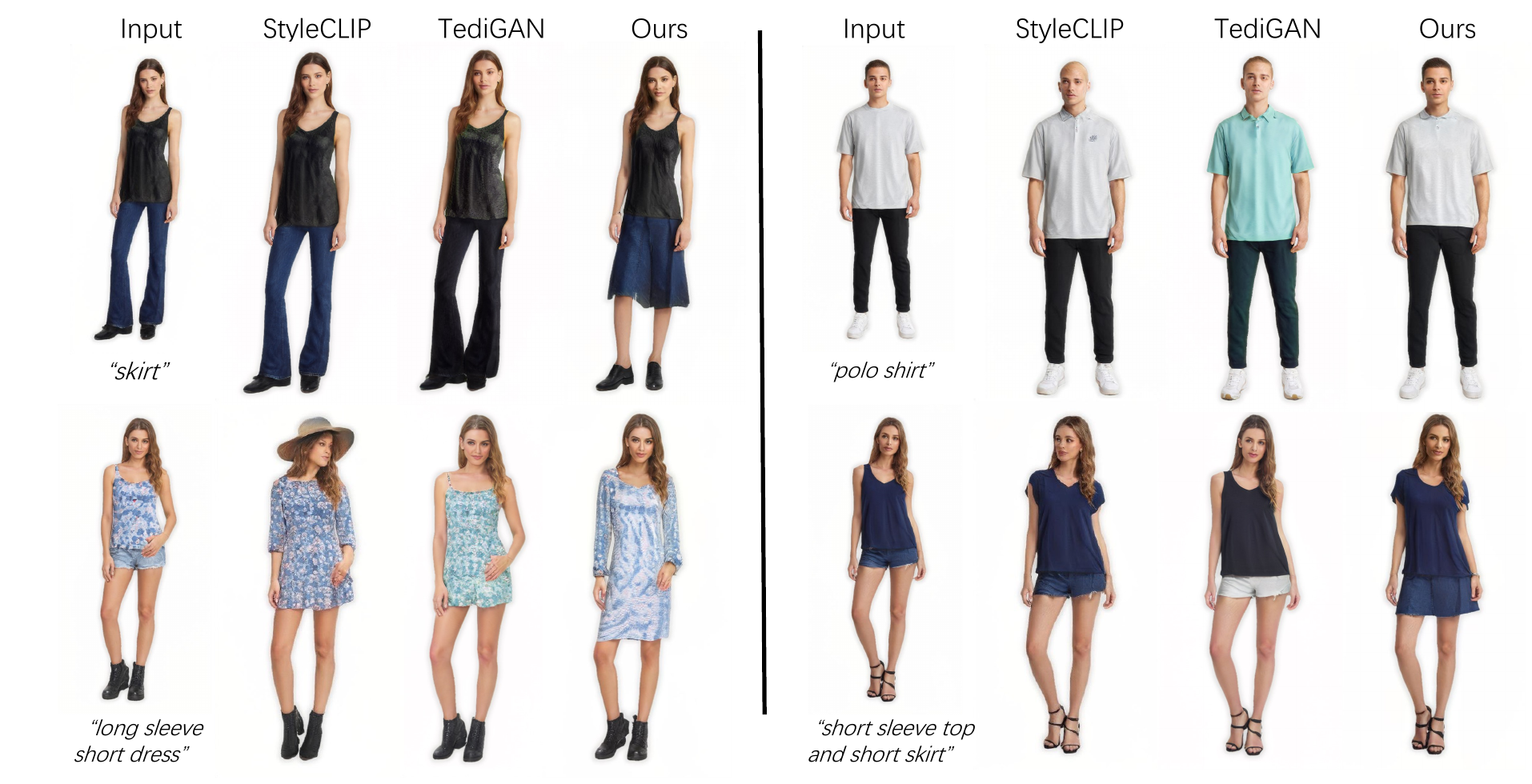}

    \vspace{-2mm}
   \caption{Qualitative comparisons on fashion type editing. We compare our method with two state-of-the-art methods: StyleCLIP\cite{patashnik2021styleclip} and TediGAN\cite{xia2021tedigan}.}
   \label{fig:com_tex}
   \vspace{-2mm}
\end{figure*}

%% file: Tables/quant.tex
\begin{table}[t]
\small
\centering
\tabcolsep7pt
 \caption{Quantitative Comparison for type editing(above) and texture transfer(below). }
  \begin{tabular}{l|c|cc}
  \hline
  &  Method & FID $\downarrow$ & Accuracy $\uparrow$ \\
  \hline
  Type &  Styleclip\cite{patashnik2021styleclip} & 90.25 & 22.25\% \\
	 &  Tedigan\cite{xia2021tedigan} & 95.44 & 15.25\% \\
		&  Ours & \textbf{69.22} & \textbf{82.75}\%  \\
\hline
\hline
&  Method & FID $\downarrow$  & LPIPS $\downarrow$ \\
\hline
Texture &  TextureGAN\cite{xian2018texturegan} & 225.28 & 0.4070 \\
		&  Texture Reformer\cite{wang2022texture} & 189.68 & 0.3687\\
		&  DiOr\cite{cui2021dressing} & 226.18 & 0.3784 \\
		&  Ours & \textbf{184.85} & \textbf{0.3257}  \\
		\hline
	\end{tabular}
\label{tab:comparison}
\end{table}

\begin{table}[t]
\small
\centering
\tabcolsep18pt
 \caption{Ablation study of our proposed Type Loss and Text Splitting method. }
  \begin{tabular}{l|c|cc}
  \hline
   Method & FID $\downarrow$ & Accuracy $\uparrow$ \\
  \hline
   
            w/o Type Loss & 172.01 & 58.75\% \\
		w/o Text Splitting  & 97.80 & 10.75\% \\
		  Ours & \textbf{69.22} & \textbf{82.75}\%  \\
\hline

	\end{tabular}
\label{tab:ablation1}
\end{table}

%% file: Sections/4.2_exp_texture.tex
\subsection{Comparisons on Fashion Texture Transfer.}
\input{Figs/comp-texture}

\paragraph{Baseline methods.}
To verify the effectiveness of our FashionTex on the task of texture transfer, three representative texture-guided generative methods are selected: 
(1) TextureGAN \cite{xian2018texturegan}.
We follow the default implementation of TextureGAN and replace the input sketch with the ground-truth segmentation map  for a fair comparison. 
(2) Dress in Order (DiOr) \cite{cui2021dressing}.
We follow the implementation of the Texture Transfer part and only change the texture input for the same comparison.
(3) Texture Reformer \cite{wang2022texture}. In this experiment, the texture image is taken as a style image, and the input image is taken as a content image. 
\paragraph{Metrics.} To quantitatively evaluate the quality of the synthesized images, we adopt two widely used evaluation metrics, which are Fréchet Inception
Distance(FID) \cite{heusel2017gans} and Learned Perceptual Image Patch Similarity (LPIPS)\cite{zhang2018unreasonable}.
\paragraph{Qualitative Results.}
The visual comparisons of texture transfer are shown in Fig.~\ref{fig:con_texture}. It is obvious that our approach achieves the best performance within all methods. 
As shown in the third row of Fig.~\ref{fig:con_texture}, the quality of the generated image by Texture Reformer is not very bad for the solid-colored texture patch.
But for other patch styles, like lines, its performance looks poor. 
As for DiOr, the visual results looks much blurry for all texture cases. The reason is that their pre-train model is overfitted, and we can not get sharp outcomes for all test samples when we change the texture patches by ourselves. 

\vspace{-1mm}
\paragraph{Quantitative Results.}
Table~\ref{tab:comparison} reports the quantitative comparisons between our FashionTex and the baselines. The advantages of our approach are obviously shown in FID and LPIPS scores, which confirms the strengths of our model on texture transfer.

%% file: Figs/comp-texture.tex




\begin{figure}[h]
    \vspace{-2mm}
  \centering
  \includegraphics[width=\linewidth]{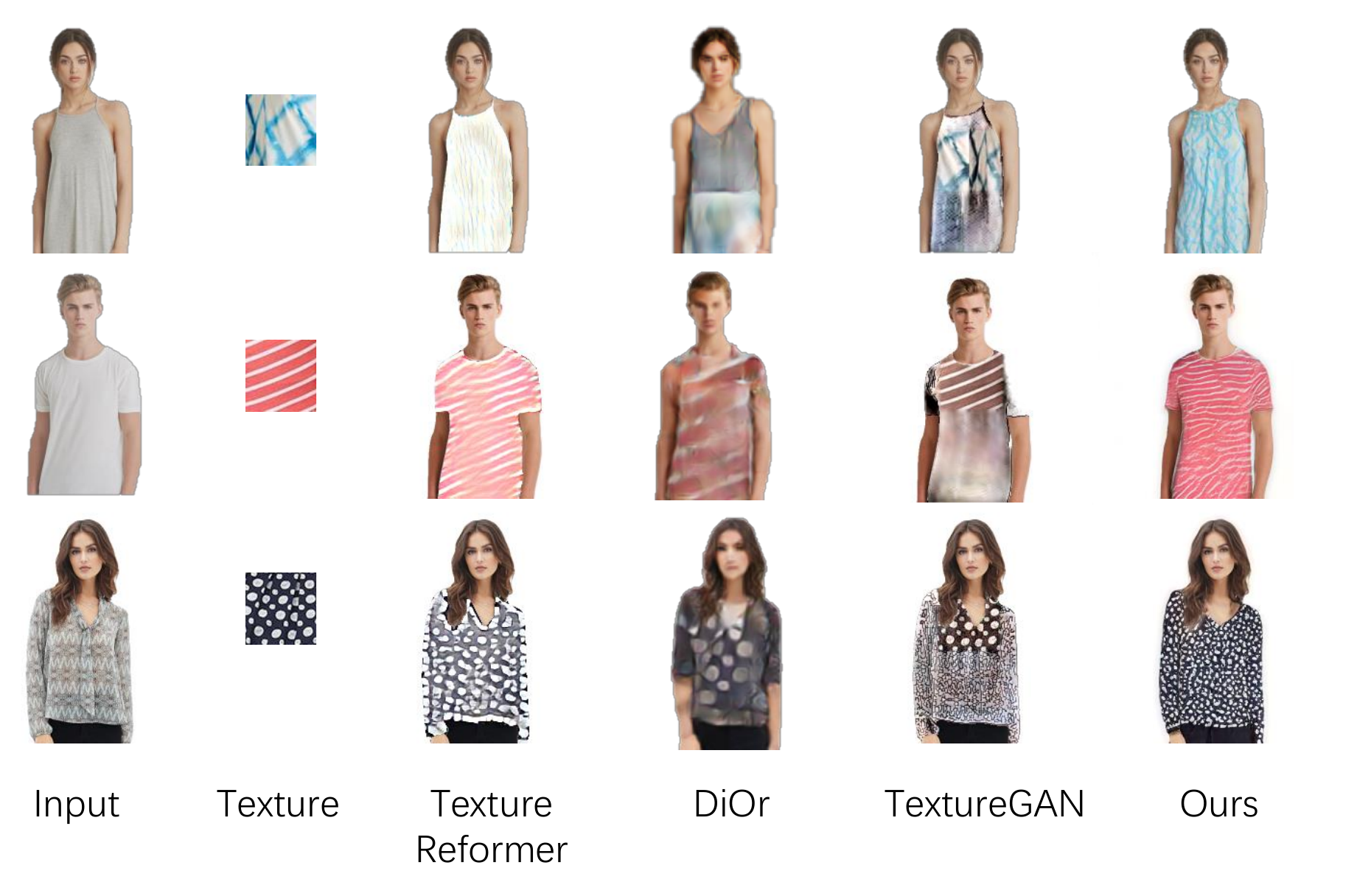}

    \vspace{-1mm}
   \caption{Qualitative comparisons on texture transfer task. We compare our method with three state-of-the-art methods: Texture Reformer\cite{wang2022texture}, DiOr\cite{cui2021dressing}, and TextureGAN\cite{xian2018texturegan}.}
    \label{fig:con_texture}
   \vspace{-2mm}
\end{figure}

%% file: Sections/4.3_ablation.tex
\subsection{Ablation Study}

\input{Figs/abla-textfortexture.tex}
\input{Figs/abla-textandid.tex}

\paragraph{The Effect of Multi-modal Interaction.} 
In our work, we use interactions in different modalities to guide the type and texture editing differently.
To demonstrate the effectiveness of this interaction method, we compare it with using only text to guide both type and texture editing. To be more specific, we replace the input texture patch reference with texture descriptions. We use three common kinds of texture, \textit{Plaid, Floral, and Striped}, with color descriptions, such as \textit{red, black, and blue}. The results can be seen in Fig.~\ref{fig:textfortexture}. Compared with text, the reference texture patch can bring more diverse and precise texture results. For example, the stripe pattern tends to be the same scale giving a text description for texture generation. 
\vspace{-1mm}
\paragraph{The Effect of Type Loss.} We compare our type loss with the naive CLIP loss (see Eq.~\ref{eq:cliploss}). The comparison results are shown in Fig.~\ref{fig:ab_typeandid}. As can be seen, using naive clip loss can achieve the change of sleeve length, but it fails to change the dress. With our type loss, the model pay more attention to the target area, which can make the image match the input condition better.
We use the same metrics as type
comparison, and the quantitative comparisons are shown in Tab.~\ref{tab:ablation1}. The first row result is the performance of naive CLIP loss.

\vspace{-1mm}
\paragraph{The Effect of ID Recovery Module.}
For the ID recovery module, we directly show the results before and after this module (see Fig.~\ref{fig:ab_typeandid}). It should be noted that the identity has been lost caused by the inversion before adding our ID recovery module, while ours achieve satisfied result.

\vspace{-1mm}
\paragraph{The Effect of Text Splitting.}
Our fashion editing module separates text prompts into descriptions of upper and lower cloth parts. This explicit separation is designed based on the structure of the human body and leads conditions better focus on the corresponding body part. And we use the same metrics to measure the editing ability. Tab.~\ref{tab:ablation1} illustrates the effectiveness of this design.

%% file: Figs/abla-textfortexture.tex
\begin{figure}[h]
    \vspace{-1mm}
  \centering
  \includegraphics[width=\linewidth]{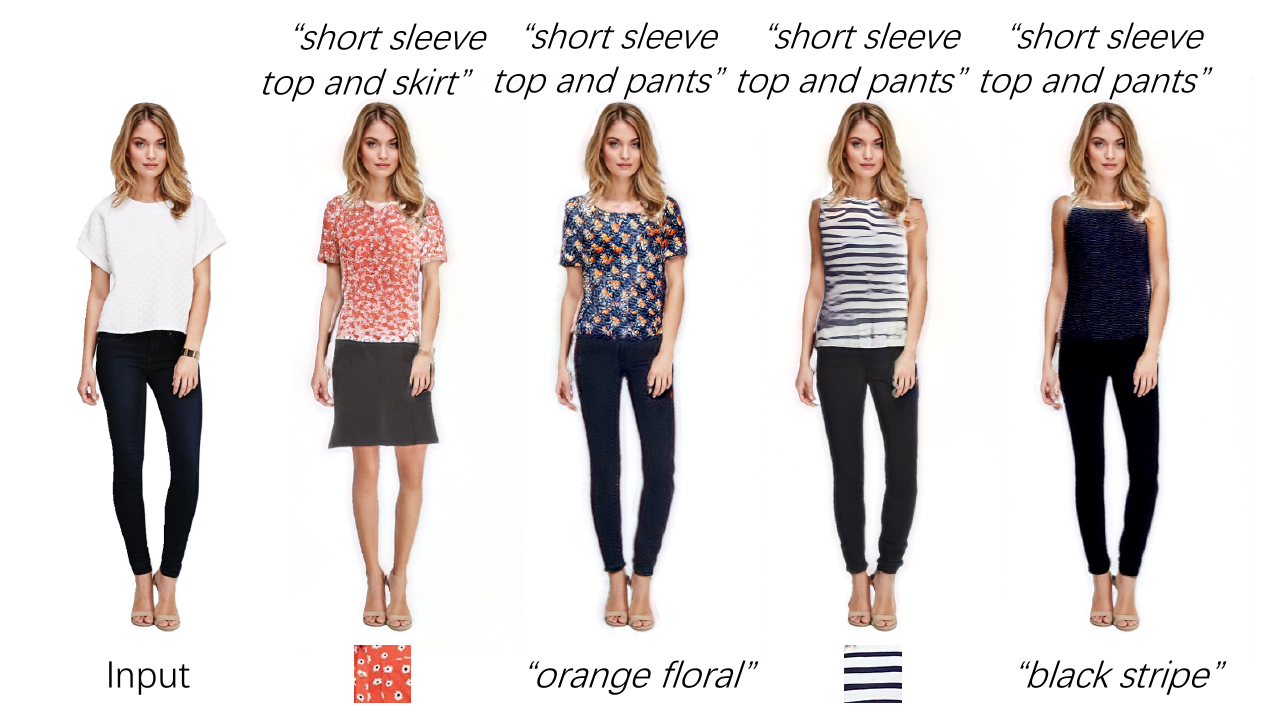}
    \vspace{-1mm}
   \caption{The results of using text prompts for both type and texture editing. The upper text description represents the target type, and below is the texture description or reference texture patch. We keep the lower cloth’s texture and only show the texture condition for upper cloth.}
   \label{fig:textfortexture}
   \vspace{-1mm}
\end{figure}

%% file: Figs/abla-textandid.tex
\begin{figure}
    \vspace{-1mm}
  \centering
  \includegraphics[width=\linewidth]{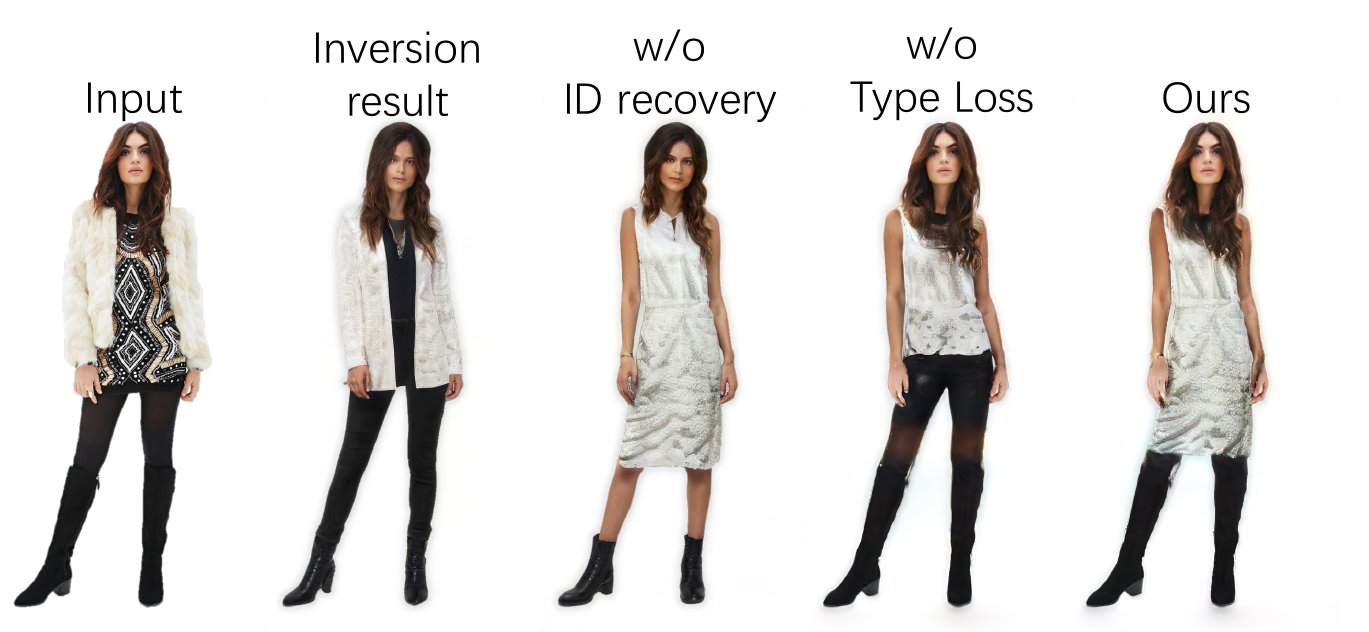}

    \vspace{-1mm}
   \caption{The effect of type Loss and ID recovery module. The input text prompt is \textit{sleeveless dress}.
}
   \label{fig:ab_typeandid}
   \vspace{-1mm}
\end{figure}

%% file: Sections/5_conclusion.tex
\section{Conclusion}
\label{sec:conclusion}
We introduce a novel and practical task of using multi-modal interactions (\textit{i.e.}, textual description and texture image patch) for virtual fashion cloth try-on. Based on a StyleGAN structure, we propose a new FashionTex pipeline that can generate high-quality results meeting the input conditions without modifying the identity of the input full-body fashion portrait. The key to our pipeline is a fashion editing module for obtaining the corresponding fashion editing displacement without pairwise training data, and an ID recovery module to preserve personal identity. Experiments have been conducted to demonstrate the effectiveness of our FashionTex. In summary, we believe that our interaction way is a powerful editing tool for virtual try-on, and hope to inspire future works along this research line.